\pdfoutput=1

\documentclass[11pt]{article}
\usepackage{listings}

\usepackage[final]{acl}

\usepackage{times}
\usepackage{tabularx}
\usepackage{latexsym}
\usepackage{amsmath} 
\usepackage{graphicx}
\usepackage[T1]{fontenc}

\usepackage[utf8]{inputenc}
\usepackage{textgreek}
\usepackage[font=small,labelfont=bf]{caption}
\usepackage{subcaption}
\usepackage{microtype}

\usepackage{inconsolata}

\usepackage{graphicx}
\usepackage{booktabs}

\title{All Entities are Not Created Equal: Examining the Long Tail for Ultra-Fine Entity Typing}

\author{Advait Deshmukh
\;\; Ashwin Umadi
\;\; Dananjay Srinivas \;\; \textbf{Maria Leonor Pacheco} \\
University of Colorado Boulder \\
\texttt{\{advait.deshmukh, ashwin.umadi, dananjay.srinivas\}@colorado.edu} }

\begin{document}

\maketitle
\begin{abstract}
Due to their capacity to acquire world knowledge from large corpora, pre-trained language models (PLMs) are extensively used in ultra-fine entity typing tasks where the space of labels is extremely large. 
In this work, we explore the limitations of the knowledge acquired by PLMs by proposing a novel heuristic to approximate the pre-training distribution of entities when the pre-training data is unknown. 
Then, we systematically demonstrate that entity-typing approaches that rely solely on the parametric knowledge of PLMs struggle significantly with entities at the long tail of the pre-training distribution, and that knowledge-infused approaches can account for some of these shortcomings. Our findings suggest that we need to go beyond PLMs to produce solutions that perform well for infrequent entities. 
\end{abstract}

\renewcommand{\thefootnote}{\fnsymbol{footnote}}

\section{Introduction}

Ultra-fine entity typing (UFET) is the task of inferring the type of an entity with high specificity~\cite{choi-etal-2018-ultra}.
For example, in the sentence \textit{``Barack Obama attended Biden's presidential inauguration.''}, the entity \texttt{``Barack Obama''} would have a coarse entity-type of \texttt{``person''}, but a more specific, ultra-fine entity type such as \texttt{``politician''}, \texttt{``democrat''} or \texttt{``ex-president''}. 
Previous work on UFET relies mostly on pre-trained language models (PLMs) to elicit entity types~\cite{li2022ultra,li2023ultrafineentitytypingprior,dai-etal-2021-ultra,ijcai2022pan}. 
The common approach is to position the UFET task close to the learning objective of the PLM. 
For example, \citet{ijcai2022pan} exploit the masked language modeling objective by appending an entity-mention and mask to a sentence, allowing BERT to fill in the mask to elicit a type (e.g., \texttt{Barack Obama attended Biden's presidential inauguration, Barack Obama is a [MASK]}). 
Such approaches are attractive because they benefit from the large amount of pre-training data the PLM has seen to make common associations, thus helping them determine the most likely type of entity based on the context provided. 

Albeit successful, PLMs are limited by the data that they have been exposed to, so their reliability can be affected when working with infrequently used language constructions such as rarely/never seen entities. Since PLMs rarely encounter these entities, they have fewer opportunities to capture knowledge about them compared to frequent entities.  
Past work has mostly overlooked this issue, assuming that the embedding space will be enough to capture similarities between rare and less rare entities and generalize across them.
Studies that have looked at infrequent entities usually characterize them in terms of their availability in task-specific training data~\cite{choi-etal-2018-ultra, schouten-etal-2022-probing}, rather than the data used to pre-train PLMs. 


In this paper, we investigate the extent to which PLM-based methods struggle when determining the ultra-fine grained type of entities that occur infrequently in their pre-training data. This is a challenging undertaking, as the data used for pre-training PLMs are often unavailable~\cite{shi2024detecting}. To address this challenge, we set out to answer the following research questions:

\textbf{RQ1: Do internet search hits provide a good proxy for estimating relative entity frequency?} To answer this question, we follow the conventional wisdom that modern PLMs have been trained on ``all of the Internet'', and we estimate the frequency of an entity by querying the Google search API to obtain the number of documents containing that entity. We validate this proxy by correlating it with real-world datasets known to be used in PLM pre-training, acknowledging that such disclosures are limited to only a few models.

\textbf{RQ2: Do internal model representations reflect the entity frequencies in the pre-training data?} To answer this question, we measure the correlation between the number of Google search hits obtained for an entity and the model's probability of eliciting that entity in numerous contexts. We perform this test across various PLMs, including both masked LMs and causal LMs, and find a strong correlation across the board. 

\textbf{RQ3: Do PLMs struggle to type entities that are in the long tail of the pre-training distribution?} Finally, we design a benchmark to answer this question.\footnote{Data and Code available at: \url{https://github.com/blast-cu/All-Entities-are-Not-Created-Equal}}
To do this, we divide entities in various UFET datasets into bins based on their resulting frequencies from our Google search. Our hypothesis is that the bins consisting of the least frequent entities ``in the wild'' will be much harder to predict than the rest. We test this assertion by looking at the performance of two types of published approaches for UFET; PLM-only approaches, which we expect to struggle at the long tail of the pre-training distribution, and knowledge-infused approaches, which we expect can level the playing field across frequencies by leveraging external sources of information. Our results confirm our hypothesis, suggesting that we need to go beyond PLMs to produce solutions that perform well for rare, new, or infrequent entities.

\section{Related Work}



UFET was proposed to generate free-form noun phrases that appropriately describe the type of a target entity~\cite{choi-etal-2018-ultra}. 
UFET has found several downstream applications, such as coreference resolution \cite{durrett-klein-2014-joint}, entity linking \cite{Onoe_Durrett_2020} and relation extraction \cite{yaghoobzadeh-etal-2017-noise}. 

\textbf{PLM solutions for UFET.} PLMs
have been shown to capture world knowledge in their parameter space~\cite{roberts-etal-2020-much,10.1162/tacl_a_00324}. This ability has allowed them to perform well on `fill-in-the-blank' problems, where the goal is to elicit an answer to a query based on the context provided. 
Many UFET approaches have capitalized on this ability; \citet{dai-etal-2021-ultra} looked at introducing Hearst patterns with \texttt{[MASK]} tokens to describe entity types using BERT, 
while \citet{ijcai2022pan} primed BERT to produce an ultra-fine type by appending the entity mention and a \texttt{[MASK]} token to a sentence. While these methods report some of the best metrics for UFET, they do not explore how their approaches fare on infrequent, rare entities that PLMs may not have been sufficiently exposed to. In this work, we explore the effect of entity representation strength on a variety of PLMs in order to determine whether rarer entities cause significant issues for PLM-based approaches. 


\textbf{Knowledge Infused UFET.} The UFET dataset has $9$ general, $121$ fine, and $10,201$ ultra-fine types. While a large type vocabulary and the scarcity of annotated examples per type make this task especially challenging, type labels often consist of rich semantics. \citet{li2022ultra} leverage the type semantics and formulate the task as an NLI problem. Others have exploited the dependencies between labels \cite{liu-etal-2021-fine} and hierarchies within types \cite{onoe-etal-2021-modeling} to supplement the PLM objectives. Such techniques often perform significantly better than PLM-only approaches.

\textbf{Approaches to Estimate Entity Frequencies in Large Corpora.} 
While there have been efforts to efficiently index large pre-training corpora to better estimate entity frequencies~\cite{Liu2024InfiniGram, xu2025infinigramminiexactngram}, we argue that there are two clear limitations with this approach: (1) the assumption that training data for models is always available and (2) the need to re-index datasets as they are updated over time. Therefore, we need simple proxies that can approximate the entity distribution ``seen'' by PLMs without direct access to their training data. 

\section{Experimental Design}\label{sec:design}

Our study is composed of three experiments. First, we evaluate a proxy to establish the relative frequency of entities in pre-training data using Search Engine API indexing. Then, we examine whether parametric representations match these frequencies by correlating the likelihood of a PLM suggesting a target entity to how the term is indexed on the Web. Lastly, we select seven models to perform ultra-fine entity typing; three based on PLM objectives, three that leverage some additional knowledge, and one simple baseline. We measure typing performance across entity groups of varying frequency across the Internet data.

\paragraph{Establishing the Long Tail (RQ1, RQ2)} The long tail of an entity distribution is characterized by a large number of entities that are rare to encounter in the real world. 
Assuming that the Internet is a fairly balanced representation of a real-world distribution, we use the Internet search hits as a proxy for frequency. We quantify the occurrence ``in the wild'' of each entity in our dataset by performing strict searches using the Google Search API. 

As most datasets used to pre-train modern PLMs come from a subset of the Internet, we hypothesize that such ``long-tail'' entities would be underrepresented in the parameter space of these models. 
To test this hypothesis, we query models and assess whether an entity is underrepresented based on its ability to be predicted or reconstructed accurately. We then measure the correlation between Internet frequencies and model-estimated probabilities. 

To perform model estimations reliably, we must balance different considerations. First, given that PLMs learn word representations from in-context examples, we need a representative set of in-context examples for each of the entities that we want to estimate. To do this, we prompt \textit{Llama3-8B-Instruct} to generate 10 different sentences that include the target entity. Second, we need to consider the training objective of the PLM when querying it for the probability of a given entity in a given context. 
To recover entity probabilities with \underline{Masked LMs}, we first replace entity tokens with \texttt{[MASK]} tokens. Then, we generate a probability distribution over all candidate tokens. 
To deal with \underline{Causal LMs}, we reframe the task as a \textit{fill-in-the-blank} problem. We provide the model with a prompt (see App \ref{sec:llama_prompt}) and compute the probabilities through a conditional generation process.

In both cases, we use the probability that the model assigns to the tokens of the target entity to represent its salience in the model's parametric knowledge. 
More details about the probability estimation process can be found in App.~\ref{sec:mask-gen}.

\begin{table}[t]
\small
  \centering
  \begin{tabularx}{\columnwidth}{@{} l c X c @{}}
    \toprule
    \textbf{Bin} &
    \shortstack{\textbf{\# of}\\ \textbf{Examples}} &
    \textbf{Representative Entity} &
    \shortstack{\textbf{Avg. \#}\\ \textbf{of Tokens}}\\
    \midrule
    1 & 301 & the Baton Rouge police chief and the serial murder task force & 11.63\\
    \midrule
    2 & 301 & Left fielder Carl Crawford & 4.35\\
    \midrule
    3 & 300 & The Polish government & 2.67\\
    \midrule
    4 & 1095 & the film & 1.18\\
    \bottomrule
  \end{tabularx}
  \caption{Distribution of entities across UFET test bins}
  \label{table:rep-entities}
\end{table}

\begin{figure}[t]
    \centering
    \includegraphics[width=0.85\columnwidth]{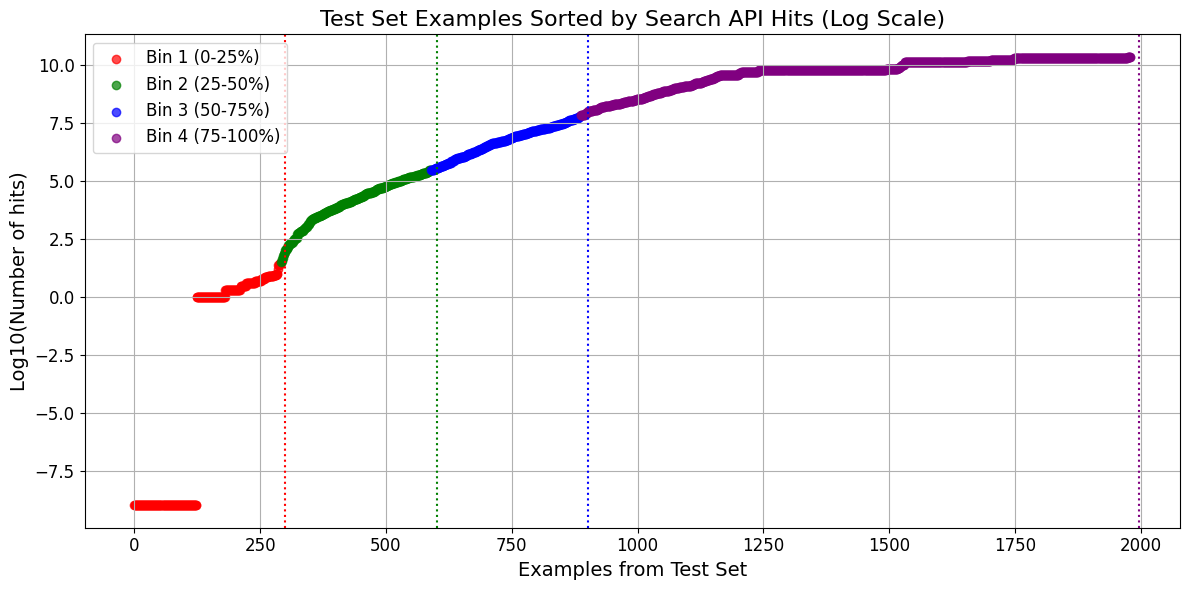}
    \caption{Entity distribution across \textbf{UFET} test bins}
    \label{fig:bins_visualized}
\end{figure}

\paragraph{Measuring Impact on Typing (RQ3)} 
We examine how the real-world distribution of entities affects the performance of entity-typing models. To define rare entities, we calculate frequency scores for all target UFET entities using the Google Custom Search API and group them into four bins based on quartiles. Bin $1$ includes the rarest entities, while Bin $4$ contains the most frequent; based on their occurrence on the Internet. 
Representative examples are presented in Tab. \ref{table:rep-entities}. The entity distribution across the bins is visualized in Fig. \ref{fig:bins_visualized}. 

We select seven representative models to test against the bin splits. 
The models can be roughly divided into two categories. (1) Naive \underline{PLM-based} approaches that rely on PLM objectives such as MLM (BERT family of models) or Causal LM (Llama3 and Qwen3) to predict types. We include implementation details in App.~\ref{sec:ufet-baselines}. (2) \underline{Knowledge-infused} approaches that exploit additional information embedded within or relevant to the type labels. LITE \cite{li2022ultra} formulates entity typing as an NLI problem. LRN \cite{liu-etal-2021-fine} exploits intrinsic and extrinsic dependencies between label types. Box4Types \cite{onoe-etal-2021-modeling} relies on box embeddings to capture hierarchies of types. More details about these systems can be found in App~\ref{sec:ufet-baselines}. For completeness, we also benchmark an LSTM model. This is a supervised model built on top of pre-trained word embeddings, which consider similar representation objectives to those of PLMs (word co-occurrence).


We use the crowd-annotated portion of the UFET dataset \cite{choi-etal-2018-ultra} for our experiments. This dataset contains entity mentions with their surrounding context and the ultra-fine types associated with them. The dataset of $5{,}994$ tuples is divided into train/test/dev splits, each containing $1{,}998$ tuples. We use OntoNotes \cite{gillick2016contextdependentfinegrainedentitytype} as a secondary dataset, which has a train/dev/test split of $250$k/$2$k/$9$k examples.
\section{Evaluation and Discussion}



\begin{figure}[t]                    
  \centering
  \begin{subfigure}[t]{0.85\columnwidth}
    \centering
    \includegraphics[width=\linewidth]{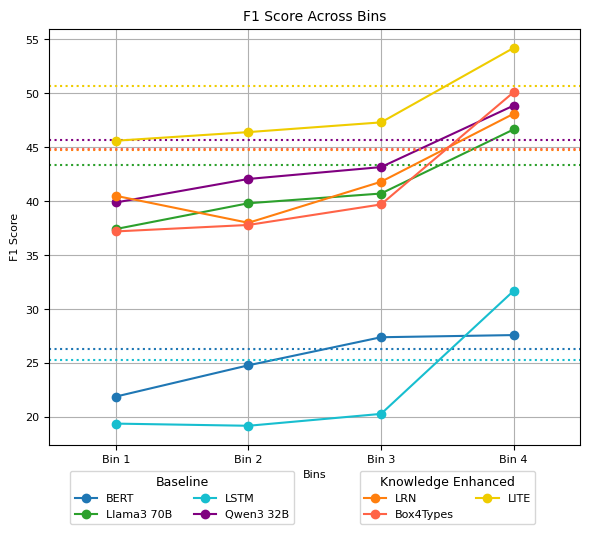}
    \caption{UFET}
    
  \end{subfigure}\hfill
  \begin{subfigure}[t]{0.85\columnwidth}
    \centering
    \includegraphics[width=\linewidth]{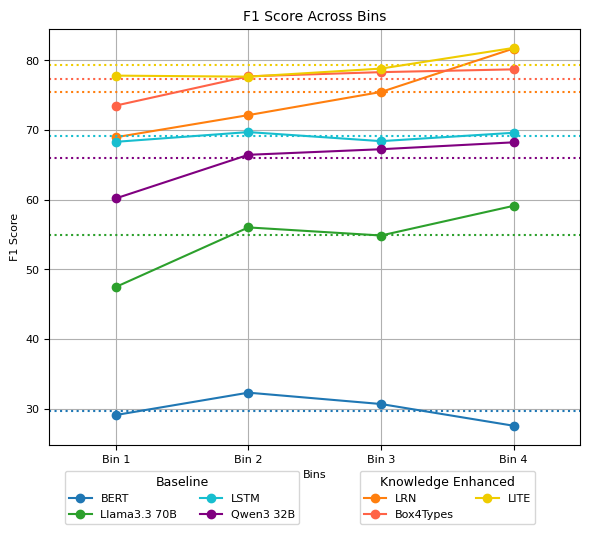}
    \caption{OntoNotes}
    \label{fig:ontonotes_bins}
  \end{subfigure}
  
  \caption{Baseline vs. Knowledge-enhanced Performance across test bins}
 \label{fig:baseline_vs_knowledge}
  

\end{figure}

\begin{table}[t]
\small
\centering
\begin{tabular}{@{}ccc@{}}
\toprule
\textbf{Training Corpus} & \textbf{Spearman Correlation} \\ \midrule
BookCorpus              &  0.583                  \\
C4-train                &  0.957                   \\
Pile-train              &  0.954                  \\
RedPajama               &  0.957                  \\
Dolma-v1.7              &  0.961                  \\ \bottomrule
\end{tabular}
\caption{Spearman correlation coefficient of \textbf{UFET} entity frequencies as estimated by search API vs counts from actual pre-training corpora}
\label{tab:api_vs_corpora}
\end{table}

\paragraph{Long Tail Analysis (RQ1, RQ2)}
We compare average PLM probability estimates with the number of hits the Search Engine API has for the target entity. We do this for three PLMs with both MLM and Causal LM learning objectives: BERT~\cite{devlin-etal-2019-bert}, BART~\cite{lewis-etal-2020-bart} and Qwen3~\cite{yang2025qwen3technicalreport}. We calculate the probability assigned by the PLMs per in-context example using the method described in Section \ref{sec:design}, and take the average across all sentences for its respective entity. We perform a correlation analysis between PLM probability estimates and the entity frequencies obtained from the Google Search API and observe high Spearman correlation coefficients (0.885 for BERT, 0.716 for BART and 0.897 for Qwen3). To visualize this, we plot the hits from the Search Engine API against the average probability for the entity obtained by each model in App.~\ref{sec:scatter_plots}. 
The high correlation between the PLM probability estimates and the number of API hits supports our hypothesis: entities that occur more/less frequently in the real world are more/less salient in PLMs.




\textit{Temporal shifts.} We recognize that the Internet is constantly evolving and that temporal dynamics could potentially alter the distribution of entities. For this reason, we performed our analysis with API data capped at 2018 and at 2024, and found the results to be consistent over time, with minor changes in correlation coefficients (See App. \ref{sec:temporal_dynamics}).

\textit{Search API vs. real pre-training datasets.} We test the validity of using Internet counts to estimate the pre-training distribution by performing a correlation study between search API hits and real-world pre-training datasets such as BookCorpus \cite{Zhu_2015_ICCV}, C4 \cite{JMLR:v21:20-074}, Pile \cite{gao2020pile, biderman2022datasheet}, RedPajama \cite{NEURIPS2024_d3449733}, Dolma-v1.7 \cite{soldaini-etal-2024-dolma}. We estimate the entity counts for BookCorpus by performing strict searches. For Pile, we rely on the Infini-gram-mini hosted API \cite{xu2025infinigramminiexactngram}. For the remaining datasets (C4, RedPajama, Dolma); we use the Infini-gram hosted API \cite{Liu2024InfiniGram}. N-gram counts in Infini-gram and Infini-gram-mini are case-sensitive and therefore noisy. Regardless, we see high spearman coefficients (>0.9) for all of them. While the correlation coefficient for BookCorpus is significantly lower, we find that the largest contributing factor is its much smaller scale. As a result, many entities are not present in BookCorpus. When restricting only to entities that are present in BookCorpus, we find that the correlation jumps as high as 0.88.

For completeness, we also report the Spearman correlation coefficients between the frequency of UFET entities in different pre-training datasets and the average entity recovery probability given surrounding context (see App. \ref{sec:datasets-vs-api}). 




\begin{figure}[t]
    \centering
    \includegraphics[width=0.85\linewidth]{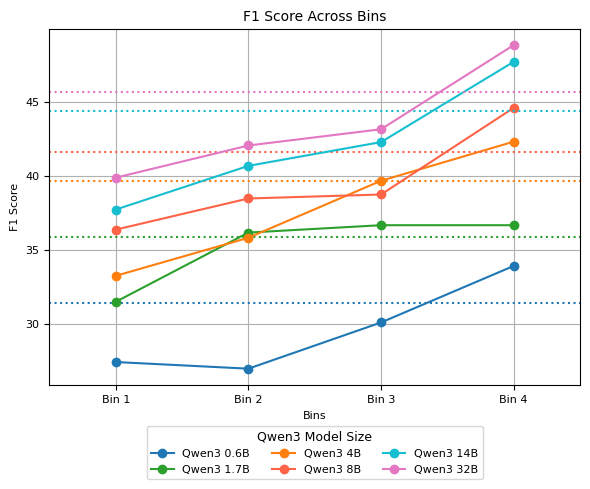}
    \caption{Effect of scaling on performance across \textbf{UFET} bins}
    \label{fig:qwen_ufet}
\end{figure}

\paragraph{Typing Performance (RQ3)}
We benchmark the seven models introduced in Sec.~\ref{sec:design}: UFET-LSTM \cite{choi-etal-2018-ultra}, few-shot Llama3 \cite{dubey2024llama3herdmodels}/Qwen3 \cite{yang2025qwen3technicalreport}, BERT (\underline{PLM-based}), and LITE \cite{li2022ultra}, LRN \cite{liu-etal-2021-fine} and Box4Types \cite{onoe-etal-2021-modeling} (\underline{Knowledge-infused}), and plot results in Fig.~\ref{fig:baseline_vs_knowledge}. 


We find that most of the examined approaches perform better in determining the ultra-fine entity type when the target entity is more frequent (Bin$4$), and worse when the entity is rarer (Bin$1$). 
We note that most PLM approaches show a major decline in performance as we go down the bins (or under-performing completely, as in the case of BERT for OntoNotes). This performance decline is most underlined when moving from Bin 4 to Bin 3. Looking at the knowledge-infused approaches, we find that the enrichment from auxiliary tasks or label dependencies helps overcome the overreliance on learned entity representations for Bin 1. LITE performs the best, beating the rest by achieving an $F1$ score of $45.6$ on infrequent entities and $54.2$ on frequent entities for UFET, and an $F1$ score of $77.8$ on infrequent entities and $81.8$ on frequent entities for OntoNotes. This further highlights the advantage of the auxiliary task (NLI), which allows the model to transfer more knowledge than approaches that only rely on the PLMs parameters. 

\begin{figure}[!t]  
  \centering

  \begin{subfigure}[t]{0.85\columnwidth}
    \centering
    \includegraphics[width=\linewidth]{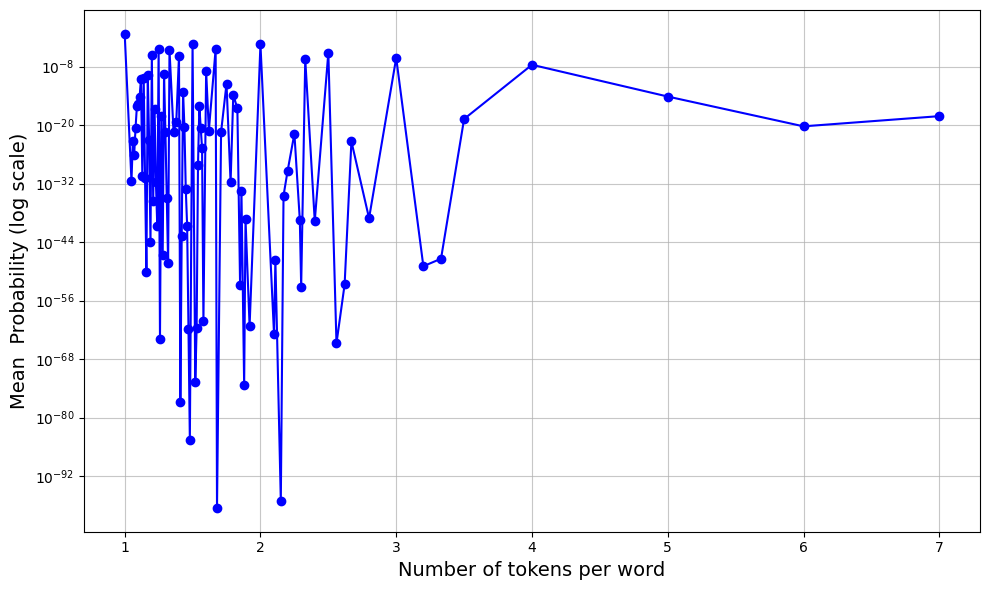}
    \caption{BERT-Base-Uncased}
    \label{fig:bert_tokens_vs_subwords}
  \end{subfigure}\hfill       

  \begin{subfigure}[t]{0.85\columnwidth}
    \centering
    \includegraphics[width=\linewidth]{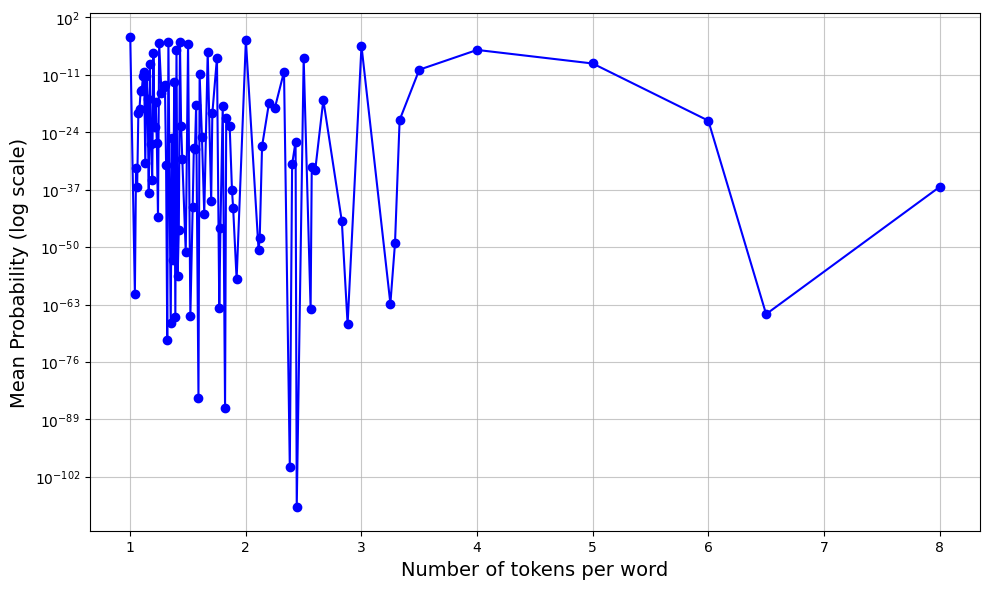}
    \caption{BART-Large}
    \label{fig:bart_tokens_vs_subwords}
  \end{subfigure}\hfill

  \begin{subfigure}[t]{0.85\columnwidth}
    \centering
    \includegraphics[width=\linewidth]{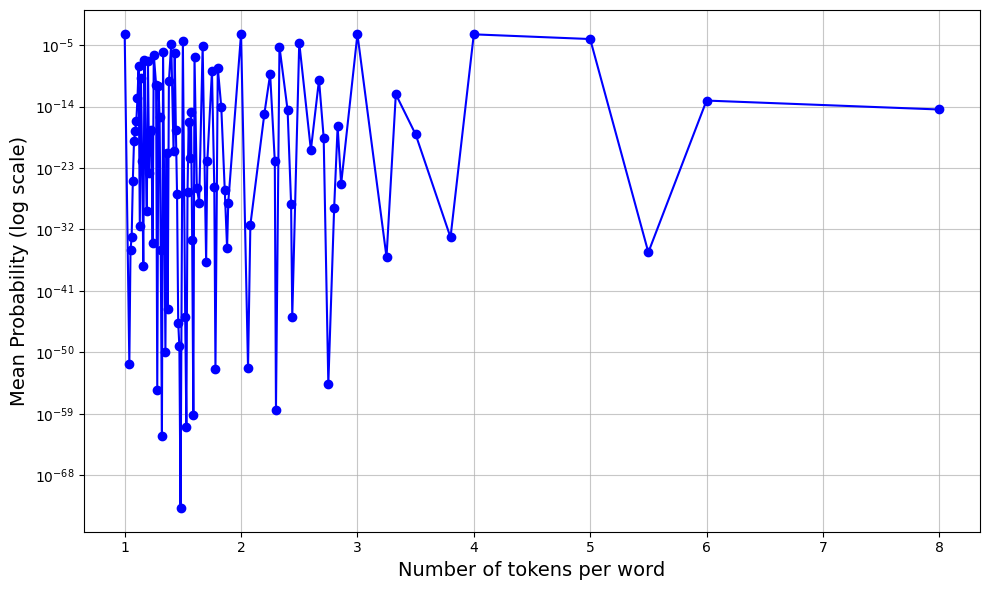}
    \caption{Qwen3-0.6B}
    \label{fig:qwen_tokens_vs_subwords}
  \end{subfigure}

  \caption{Average \textbf{UFET} entity recovery probability versus average number of tokens per word for three model tokenizers}
  \label{fig:tokens_vs_subwords}

\end{figure}

\textit{Effects of tokenizers.} If a specific word is unknown to a PLM, it follows that the tokenizer will split it into multiple tokens. For this reason, we investigate whether words being split into multiple tokens causes a degradation in the probability of recovering an entity by looking at the splitting ratio (number of tokens / number of words). While we see a marked difference between a ratio of $1$ and a ratio $\neq 1$, our results do not suggest any incremental effect afterwards (see Fig. \ref{fig:tokens_vs_subwords}). This suggests that our API Search proxy is a better, more nuanced indicator of entity rarity than the PLM tokenizer. 

\textit{Other levels of granularity.} We look at performance across frequency bins at varying entity type granularity (i.e., fine, coarse) and find the same trend of performance decay across buckets within each granularity level (see App. ~\ref{sec:fine_grained_evaluation}). 

\textit{Effect of Scaling.}\label{sec:scaling-effect} To understand how the size of the LM affects the typing task across bins, we evaluate Qwen3 models of different sizes (Fig. \ref{fig:qwen_ufet}). We chose Qwen3 given the availability of a large range of model sizes of the same family. We find that regardless of model size, the effect of entity frequency in performance degradation is still observed. We include results for Llama3 in App. \ref{sec:ufet-baselines}.

\section{Conclusion and Future Work}
We showed the effectiveness of internet search API hits as a proxy for entity frequency in large pre-training corpora (RQ1). 
We also showed that, as expected, this distribution significantly affects entity representations for different PLMs (RQ2).
Finally, we measured the performance of several PLM-based and knowledge-infused entity typing systems on entities with different frequencies and found that all models performed worse on rarer/less-probable entities (RQ3). However, we found that all knowledge-infused systems are considerably more robust to frequency shifts than PLM-based approaches. 
Our results show that for PLMs to fulfill their promise for long-tail entity typing, we need better strategies to inject knowledge about rare entities into PLMs by using external resources and other forms of domain knowledge. 

\newpage
\section*{Limitations}

Due to limited resources, we used quantized versions of the models when studying the impact of scaling on entity typing systems in Sec. \ref{sec:scaling-effect}. Previous work has shown that techniques such as quantization can impact the performance of LLMs, especially in low-resource settings \cite{diddee-etal-2022-brittle}. We acknowledge that quantization may be further exacerbating the general struggle that PLMs have with entities at the long-tail of the distribution.

\section*{Ethical Considerations}

To the best of our knowledge, this work does not incur any violation of the code of ethics. 
We used models that were Open Sourced by their authors, with code available online. 
All the information required to replicate our experiment is provided in the paper. 
We use Llama3/Qwen3, large language models whose weights may be updated by the model's creator.  
In such a case, we caution that some results may not be reproducible exactly, but believe that our findings will still hold. 

In the interest of space, we moved some plots and details to the appendix.


\bibliography{custom}


\appendix
\section{Analysis of UFET entities in pre-training datasets} \label{sec:datasets-vs-api}

We report the spearman correlation values for UFET Entity counts as obtained in pre-training datasets and their recovery probabilities in Tab. \ref{tab:api_vs_corpora}.


\begin{table*}[t]
\small
\centering
\begin{tabular}{@{}ccccccc@{}}
\toprule
\textbf{Model}              & \textbf{BookCorpus} & \textbf{Pile-train} & \textbf{C4-train} & \textbf{Dolma-v1.7} & \textbf{RedPajama}   \\ \midrule
Bert-base-uncased           & 0.571               &  0.883              & 0.886             & 0.891               & 0.886                \\
Bart-large                  & 0.359               &  0.705              & 0.707             & 0.715               & 0.708                \\
Qwen3-0.6B                  & 0.514               &  0.895              & 0.897             & 0.902               & 0.901                \\
\bottomrule
\end{tabular}
\caption{Spearman correlation coefficient of \textbf{UFET} entity frequencies in pretrained datasets vs Average  entity recovery probability given surrounding context}
\label{tab:bookcorpus_vs_models}
\end{table*}

\section{Temporal dynamics of search API hits} \label{sec:temporal_dynamics}
We use the Google Search API to approximate the distribution of entity frequencies that models have seen during training. While convenient, this approach may potentially ignore the temporal changes that might occur in the distributions of such entities. This is especially important as models we discuss in our work (BERT, BART and Qwen3) have been trained at different points within the last decade. To ensure that these dynamics do not significantly impact our claims, we compare the correlation between the LM predictions and the API data capped at 2018 and 2024. We find results to be largely consistent across these two time periods (See Tab. \ref{tab:api_vs_models}). On further examination, we find that between 2018 and 2024 only 39 entities from the test set (<2\%) change their bin classification. For our main results, we rank entities using the 2024 results.

\begin{table}[t]
\small
\centering
\begin{tabular}{@{}ccc@{}}
\toprule
\textbf{Model}             & \textbf{API till 2018} & \textbf{API till 2024}      \\ \midrule
Bert-base-uncased          & 0.883                  & 0.885                       \\
Bart-large                 & 0.711                  & 0.716                       \\
Qwen3-0.6B                 & 0.895                  & 0.897                       \\ \bottomrule
\end{tabular}
\caption{Spearman correlation coefficient of API entity frequencies vs Average \textbf{UFET} entity recovery probability given surrounding context (API results capped at 2018 and 2024)}
\label{tab:api_vs_models}
\end{table}

\section{UFET test bin distribution} \label{sec:UFET_bins_visualized}

\begin{figure}[t]
    \centering
    \includegraphics[width=0.9\columnwidth]{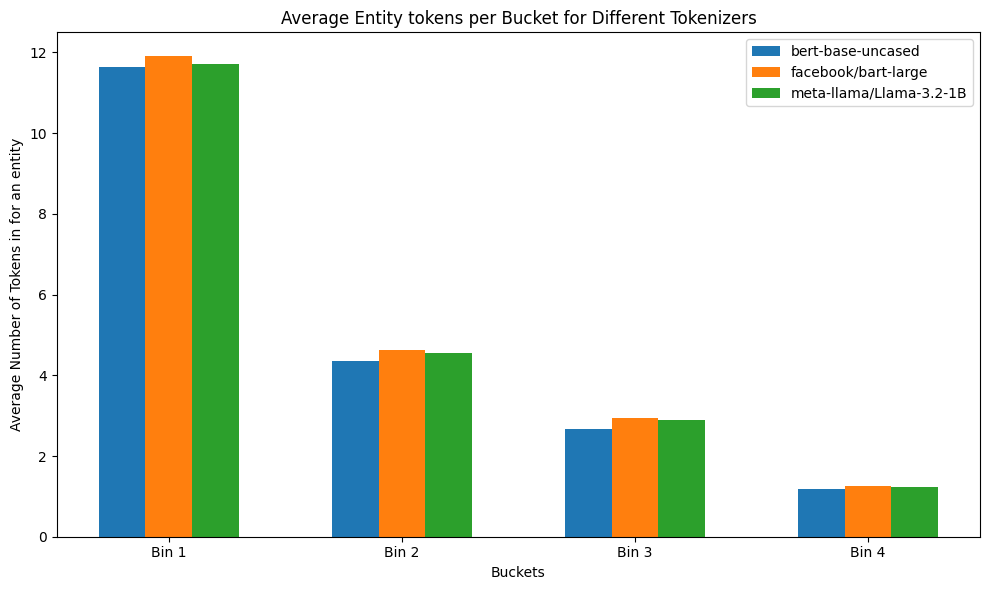}
    \caption{Average number of tokens for \textbf{UFET} test bins}
    \label{fig:tokens_per_bin}
\end{figure}

To better visualize the distribution of entities across  bins, we plot the log of API hits in Fig. \ref{fig:bins_visualized}. We also compare the average number of tokens as obtained with different tokenizers for each of our test bins in Fig. \ref{fig:tokens_per_bin}.

\section{Method for Generating Masks and Calculating Entity Probability } \label{sec:mask-gen}

An entity can be comprised of a single token or a multi-token phrase. 
For multi-token entities we employ a conditional generation approach where we generate the entity sequentially, one token at a time, moving from left to right. The probability of the entire entity is determined by product of the conditional probability of each token \( t_i \) conditioned on all preceding tokens and the surrounding context \( C \), where \( i \) represents the position of the token in the sequence and \( n \) is the total number of tokens comprising the entity.

\begin{equation}
\small
    P(t_1, t_2, \ldots, t_n \mid C) = \prod_{i=1}^{n} P\left(t_i \,\bigg|\, t_1, t_2, \ldots, t_{i-1}, C\right)
\end{equation}

We use this approach to benchmark three models with comparable sizes and distinct pre-training objectives.


\textbf{BERT (MLM):}
        To recover an entity using BERT (bert-base-uncased), we first replace the entity with an equal number of \texttt{[MASK]} tokens. Then we calculate the probability the entity being recovered as described before, replacing one \texttt{[MASK]} per iteration.

\textbf{BART (MLM):}
        For BART (bart-large) \cite{lewis-etal-2020-bart}, we take a similar approach but with only a single \texttt{<mask>} token. We progressively expand the \texttt{<mask>}, one token at a time, calculating the probability for each subsequent token until the entire entity is recovered.

\textbf{Qwen3 (Causal LM):} 
        Since Qwen3 \cite{yang2025qwen3technicalreport} is not pre-trained with an MLM objective, we reframe the task as a \textit{Fill-in-the-Blank} problem using \textit{Qwen3-0.6B}. We provide the model with a prompt (See App.~\ref{sec:llama_prompt}) and compute the probabilities through a conditional generation process, one token at a time.

\section{Prompt used to calculate the probability to recover an entity for Qwen3 }
\label{sec:llama_prompt}

    The prompt used to calculate the probability to recover an entity for Qwen3 is:

    \begin{description}
                \item[\texttt{Instruction:}] Fill in the appropriate entity that completes the sentence below.
                
                \item[\texttt{Context:}] \{sentence with the entity mention replaced by a [blank]\}
                
                \item[\texttt{Response:}] \texttt{[blank]} can be replaced with:
    \end{description}

\section{Scatter plots of Entity Recovery Probability in BERT, BART and Qwen3 against Search Engine API hits}\label{sec:scatter_plots}

\begin{figure}[!t]  
  \centering
  \begin{subfigure}[t]{0.95\columnwidth}
    \centering
    \includegraphics[width=\linewidth]{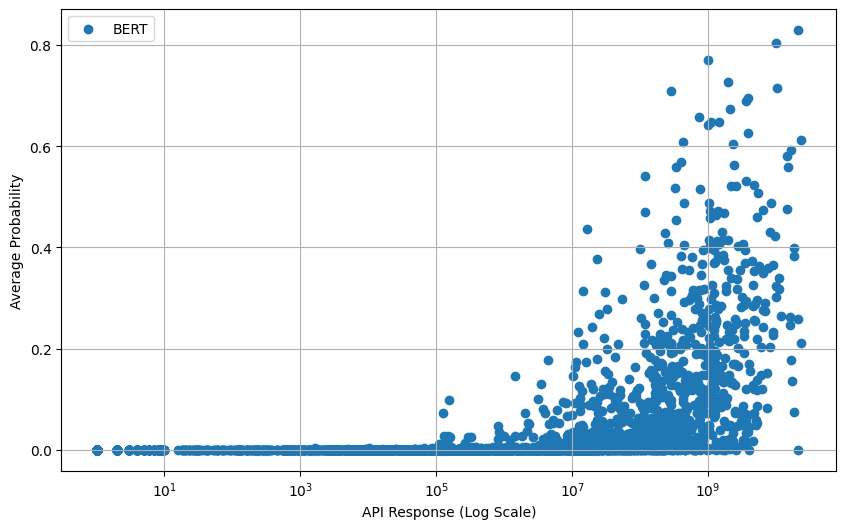}
    \caption{BERT-Base-Uncased: Average probability vs.\ Log search‐engine hits}
    \label{fig:api_vs_bert}
  \end{subfigure}\hfill%
  
  \begin{subfigure}[t]{0.95\columnwidth}
    \centering
    \includegraphics[width=\linewidth]{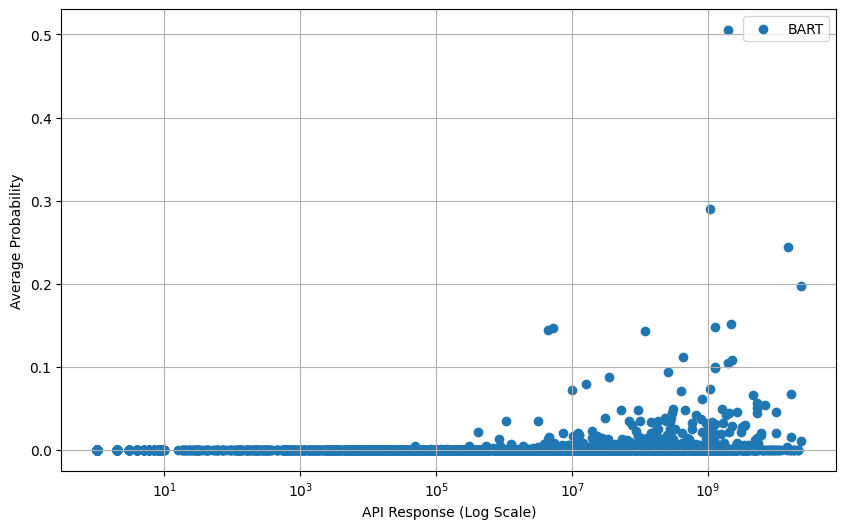}
    \caption{BART-Large: Average probability vs.\ Log search‐engine hits}
    \label{fig:api_vs_bart}
  \end{subfigure}

  \begin{subfigure}[t]{0.95\columnwidth}
    \centering
    \includegraphics[width=\linewidth]{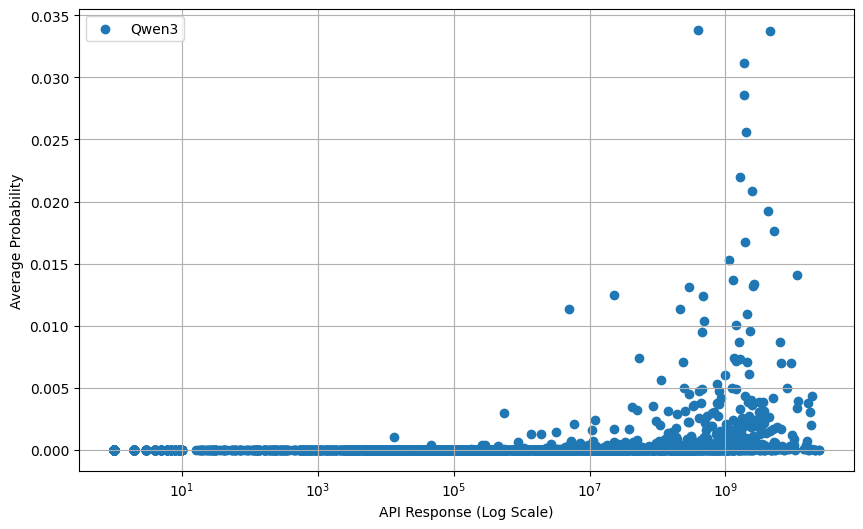}
    \caption{Qwen3-0.6B: Average probability vs.\ Log search‐engine hits}
    \label{fig:api_vs_qwen}
  \end{subfigure}\hfill%
  \caption{Average \textbf{UFET} Entity Recovery Probability vs Search API Hits}
  \label{fig:api_vs_models}
\end{figure}

To visualize the high Spearman correlation coefficient between entity recovery probabilities and search engine API hits, we plot Fig. \ref{fig:api_vs_models}.

    
\section{Models for entity typing} \label{sec:ufet-baselines}

\subsection{BERT - Baseline}
    Inspired by \cite{dai-etal-2021-ultra}, we frame the typing problem as mask prediction task for BERT (bert-base-uncased). We use hearst-like patterns ("[MASK] such as {entity mention}", "{entity mention} and any other [MASK]", "{entity mention} and some other [MASK]") and conduct experiments to find the optimal templates. The top `n' predictions for the [MASK] (with plural → singular conversion, restricted to the type vocabulary) are considered as candidate labels. we find the optimal value for `n' (number of labels) by experimenting with the development set.

    \begin{description}
        \item[\texttt{\{Hearst\}}] n = 12, F1 = 0.0661
        \item[\texttt{\{Sentence\} . \{Hearst\}}] n = 5, F1 = 0.2277
        \item[\texttt{\{Sentence\} [SEP] \{Hearst\}}] n = 5, F1 = 0.2338
        \item[\texttt{\{Hearst\} inserted in \{sentence\}}] n = 6, F1 = 0.2631
    \end{description}

\begin{figure}[t]
    \centering
    \includegraphics[width=0.9\columnwidth]{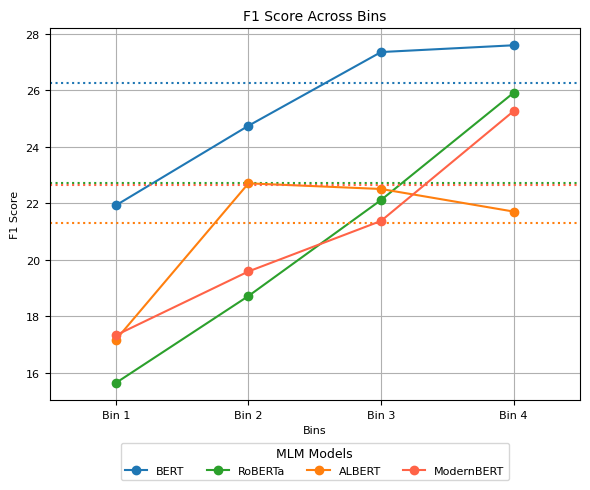}
    \caption{Evaluation of MLM models across \textbf{UFET} test bins}
    \label{fig:MLM_Models}
    \end{figure}

    We experiment with different MLM models (See Fig. \ref{fig:MLM_Models}) and present the results from the best performing setting in Tab. \ref{tab:bert_mlm}. We find that the trend is largely consistent among them, with the exception of ALBERT which converges by Bin 3.

\subsection{Llama3/Qwen3 - Baseline}

    We model the entity typing problem as a few-shot task for \textit{Llama3} and \textit{Qwen3} models to evaluate its efficacy in entity typing. We experiment with the number of examples (from the train set) in the prompts in increments of five examples. We found that the performance was optimal for 15-examples in the prompt and used that setting for the rest of our experiments.

    We use the following system prompt:
    \begin{lstlisting}
# Entity-Typing Assistant
You are a precise entity-typing assistant.  
Given a sentence in which **one entity mention is wrapped in `<ENT> ... </ENT>` tags**, produce **only** a JSON object whose single key is **"predicted_types"**.

## Guidelines
- The value must be a JSON array of strings. 
- Include all the type labels that are relevant.
- Remove duplicates and keep each type concise (ideally a short noun phrase).  
- Do not output any keys other than `"predicted_types"`.

## Input Format
- SENTENCE: The complete sentence with the target entity clearly marked with `<ENT>` tags
- ENTITY_MENTION: The target entity mention from the sentence

## Output Format
```json
{ 
    "predicted_types": ["TypeA", "TypeB", "TypeC", ...] 
}
```
    \end{lstlisting}
    Followed by examples from the train set in this format:
    \begin{lstlisting}
# Example #{i}:
- INPUT:
- SENTENCE: '{sentence}'
- ENTITY_MENTION: '{entity_mention}'

- OUTPUT:
{{\"predicted_types\": [{types}]}}   
    \end{lstlisting}
    With the input prompt as specified above, we generate the response from the model with the json schema for generation passed to the model (format parameter in ollama's client.chat method). Passing the generation schema ensures that the model adheres to the expected format and prevents malformed/incorrect json output. We filter out the generated types to match the type vocabulary for the respective dataset.

\begin{figure}[t]
        \centering
        \includegraphics[width=0.9\columnwidth]{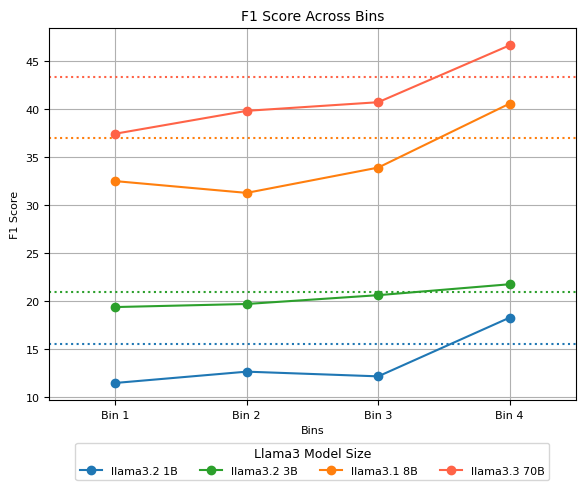}
        \caption{Effect of scaling on model performance across \textbf{UFET} test bins}
        \label{fig:llama_ufet}
    \end{figure}

    We also studied the effect of scaling on typing performance. Specifically, it was of interest to us to understand if scaling can help bridge performance discrepancies between Bin 1 and Bin 4. Although we did not find significant trends, the gains in performance made by the models generally seemingly favor buckets with higher entity frequency, exacerbating the long-tail problem. (see Fig. \ref{fig:qwen_ufet} and Fig. \ref{fig:llama_ufet})

    Note: all the models we evaluate are the 4 bit quantized models (q4\_K\_M) available through Ollama.
    
\subsection{UFET-LSTM \cite{choi-etal-2018-ultra}}
UFET-LSTM frames mention typing as predicting free-form labels (e.g., criminal, victim). Because the label set is so large and unconstrained, they adopt a neural model that represents both the mention and its context: a BiLSTM with attention for the sentence, plus a CNN with attention for the mention span. Although the authors’ reported results rely on distant supervision and crowd sourced data, our analysis uses only the crowd sourced training data from UFET. Accordingly, we train a new LSTM-based model using that dataset and present the per-bin performance in Tab. \ref{tab:baseline_LSTM}.

\subsection{LITE \cite{li2022ultra}}
LITE approaches the ultra-fine entity typing by reframing it as an NLI task: treat the original sentence (with the target entity) as a premise, and generate short textual descriptions of the entity as hypotheses using a predefined structure. A pretrained NLI model scores how strongly each description is entailed. Using a learning-to-rank objective, LITE distinguishes correct types from incorrect ones. For our experiments to find the performance across bins, we use the final results as shared by the authors on the test dataset of crowd sourced UFET. Table \ref{tab:LITE} uses these predictions while capturing its performance. For OntoNotes, we recreated the model to the best of our ability.

\subsection{Box4Types \cite{onoe-etal-2021-modeling}}
Box4Types deploys box embeddings to effectively capture the hierarchies of types. The model represents both types and entity mentions as boxes. Each mention and its
context are fed into a BERT-based model to embed that mention in our box space; essentially, the model leverages typological clues present in the surface text to hypothesize a type representation for the mention. This helps the model capture latent hierarchies better than the vector-based counterparts. We recreated the original approach described in paper and evaluated it against our bins in Tab. \ref{tab:box4types_box}. 

\subsection{Label Reasoning Network \cite{liu-etal-2021-fine}}
Label Reasoning Network sequentially reasons about fine-grained entity labels by discovering and exploiting knowledge about label dependencies entailed by the data. These implicitly and explicitly entailed dependencies provide critical information which help the model overcome limitations of baseline LM approaches. The BERT based model leverages deductive and inductive reasoning. We recreated the models (without retrieval) as described in the paper and reported the results against our bins in Tab. \ref{tab:LRN_seq}.


\section{Fine grained evaluation of the models studied} \label{sec:fine_grained_evaluation}
We look at the performance of the discussed models across bins and label granularities (Coarse, Fine, Ultra-fine) as first proposed by \cite{choi-etal-2018-ultra}. 

The trend of decline in performance between Bin 4 to Bin 1 continues into the fine grained evaluation for the models. For each level of label granularity we find clear separation in performance levels as we move between the bins, suggesting that binning provides a measure of difficulty independent of the label granularity. This provides a unique opportunity to approach the entity typing task from a new perspective.

\begin{table*}
\centering
\footnotesize
\caption{BERT MLM (bert-base-uncased)}
\resizebox{\textwidth}{!}{%
\begin{tabular}{l ccc ccc ccc ccc}
\toprule
 & \multicolumn{3}{c}{Overall} & \multicolumn{3}{c}{Coarse} & \multicolumn{3}{c}{Fine} & \multicolumn{3}{c}{Ultra-fine} \\
\cmidrule(lr){2-4} \cmidrule(lr){5-7} \cmidrule(lr){8-10} \cmidrule(lr){11-13}
Subset & P & R & F1 & P & R & F1 & P & R & F1 & P & R & F1 \\
\midrule
Full Test & 23.7 & 29.5 & 26.3 & 64.9 & 43.0 & 51.7 & 34.2 & 40.7 & 37.1 & 16.8 & 22.5 & 19.2 \\
Bin 1       & 19.2 & 25.5 & 21.9 & 59.0 & 41.8 & 48.9 & 22.4 & 34.3 & 27.1 & 14.1 & 19.0 & 16.2 \\
Bin 2       & 22.5 & 27.7 & 24.8 & 57.6 & 34.2 & 43.0 & 31.7 & 43.2 & 36.6 & 17.3 & 23.2 & 19.8 \\
Bin 3       & 24.6 & 31.0 & 27.4 & 60.1 & 41.0 & 48.7 & 34.4 & 42.1 & 37.9 & 18.8 & 25.1 & 21.5 \\
Bin 4       & 25.1 & 30.8 & 27.6 & 69.1 & 45.8 & 55.1 & 38.2 & 41.1 & 39.6 & 16.8 & 22.5 & 19.2 \\
\bottomrule
\end{tabular}%
}
\label{tab:bert_mlm}
\end{table*}

\begin{table*}
\centering
\footnotesize
\caption{Llama3.3-70B}
\resizebox{\textwidth}{!}{%
\begin{tabular}{l ccc ccc ccc ccc}
\toprule
 & \multicolumn{3}{c}{Overall} & \multicolumn{3}{c}{Coarse} 
 & \multicolumn{3}{c}{Fine}   & \multicolumn{3}{c}{Ultra-fine} \\
\cmidrule(lr){2-4}\cmidrule(lr){5-7}\cmidrule(lr){8-10}\cmidrule(lr){11-13}
Subset & P(±σ) & R(±σ) & F1(±σ)
       & P(±σ) & R(±σ) & F1(±σ)
       & P(±σ) & R(±σ) & F1(±σ)
       & P(±σ) & R(±σ) & F1(±σ) \\
\midrule
Full Test      & 45.9 ± 0.1 & 41.1 ± 0.1 & 43.4 ± 0.1
         & 78.2 ± 0.3 & 68.7 ± 0.2 & 73.1 ± 0.3
         & 58.0 ± 0.3 & 56.0 ± 0.2 & 57.0 ± 0.2
         & 36.6 ± 0.2 & 30.0 ± 0.2 & 33.0 ± 0.2 \\
Bin 1  & 38.3 ± 0.3 & 36.6 ± 0.7 & 37.4 ± 0.5
         & 70.7 ± 0.5 & 54.8 ± 0.7 & 61.7 ± 0.4
         & 48.5 ± 0.7 & 44.4 ± 1.0 & 46.4 ± 0.9
         & 32.1 ± 0.3 & 30.5 ± 0.7 & 31.3 ± 0.5 \\
Bin 2  & 43.2 ± 0.3 & 36.9 ± 0.3 & 39.8 ± 0.2
         & 73.7 ± 0.8 & 50.9 ± 0.9 & 60.2 ± 0.9
         & 55.0 ± 1.0 & 54.8 ± 0.8 & 54.9 ± 0.5
         & 36.5 ± 0.7 & 30.0 ± 0.3 & 33.0 ± 0.4 \\
Bin 3  & 44.4 ± 0.2 & 37.6 ± 0.3 & 40.7 ± 0.2
         & 75.7 ± 1.0 & 59.5 ± 1.1 & 66.7 ± 1.1
         & 59.4 ± 1.3 & 53.0 ± 0.9 & 56.0 ± 1.0
         & 34.7 ± 0.4 & 28.4 ± 0.2 & 31.2 ± 0.2 \\
Bin 4  & 49.0 ± 0.2 & 44.5 ± 0.1 & 46.6 ± 0.1
         & 80.9 ± 0.3 & 78.2 ± 0.5 & 79.5 ± 0.4
         & 60.4 ± 0.1 & 59.7 ± 0.4 & 60.1 ± 0.2
         & 38.4 ± 0.2 & 30.4 ± 0.3 & 33.9 ± 0.3 \\
\bottomrule
\end{tabular}%
}
\label{tab:llama70b}
\end{table*}

\begin{table*}
\centering
\footnotesize
\caption{Qwen3-32B}
\resizebox{\textwidth}{!}{%
\begin{tabular}{l ccc ccc ccc ccc}
\toprule
 & \multicolumn{3}{c}{Overall} & \multicolumn{3}{c}{Coarse} & \multicolumn{3}{c}{Fine} & \multicolumn{3}{c}{Ultra-fine} \\
\cmidrule(lr){2-4} \cmidrule(lr){5-7} \cmidrule(lr){8-10} \cmidrule(lr){11-13}
Subset & P(±σ) & R(±σ) & F1(±σ)
       & P(±σ) & R(±σ) & F1(±σ)
       & P(±σ) & R(±σ) & F1(±σ)
       & P(±σ) & R(±σ) & F1(±σ) \\
\midrule
Full Test      & 50.6 ± 0.6 & 41.6 ± 0.6 & 45.7 ± 0.3
         & 77.6 ± 0.5 & 75.1 ± 0.9 & 76.4 ± 0.3
         & 57.6 ± 0.8 & 55.9 ± 0.9 & 56.7 ± 0.4
         & 39.5 ± 0.5 & 29.0 ± 0.5 & 33.5 ± 0.4 \\
Bin 1  & 41.1 ± 1.0 & 38.7 ± 0.8 & 39.9 ± 0.7
         & 67.0 ± 0.9 & 67.5 ± 0.8 & 67.2 ± 0.8
         & 48.0 ± 1.7 & 48.5 ± 1.5 & 48.2 ± 1.4
         & 33.2 ± 0.7 & 29.5 ± 0.9 & 31.3 ± 0.7 \\
Bin 2  & 46.1 ± 0.6 & 38.7 ± 0.6 & 42.1 ± 0.3
         & 74.6 ± 1.4 & 66.6 ± 1.1 & 70.4 ± 0.8
         & 52.5 ± 0.7 & 53.9 ± 1.9 & 53.2 ± 0.7
         & 37.8 ± 0.8 & 28.6 ± 0.3 & 32.6 ± 0.4 \\
Bin 3  & 48.4 ± 1.1 & 39.0 ± 0.6 & 43.2 ± 0.4
         & 71.7 ± 1.8 & 68.7 ± 0.6 & 70.1 ± 0.6
         & 57.6 ± 1.2 & 54.6 ± 1.4 & 56.1 ± 1.2
         & 38.5 ± 1.0 & 28.4 ± 1.0 & 32.7 ± 0.8 \\
Bin 4  & 55.0 ± 0.6 & 44.0 ± 0.7 & 48.9 ± 0.3
         & 82.2 ± 0.4 & 80.2 ± 1.0 & 81.2 ± 0.5
         & 61.2 ± 0.9 & 58.3 ± 1.4 & 59.7 ± 0.7
         & 42.1 ± 0.6 & 29.1 ± 0.7 & 34.4 ± 0.4 \\
\bottomrule
\end{tabular}%
}
\label{tab:qwen3_32b}
\end{table*}

\begin{table*}
\centering
\footnotesize
\caption{LSTM}
\resizebox{\textwidth}{!}{%
\begin{tabular}{l ccc ccc ccc ccc}
\toprule
 & \multicolumn{3}{c}{Overall} & \multicolumn{3}{c}{Coarse} & \multicolumn{3}{c}{Fine} & \multicolumn{3}{c}{Ultra-fine} \\
\cmidrule(lr){2-4} \cmidrule(lr){5-7} \cmidrule(lr){8-10} \cmidrule(lr){11-13}
Subset & P & R & F1 & P & R & F1 & P & R & F1 & P & R & F1 \\
\midrule
Full test  & 41.7 & 18.2 & 25.3 & 57.3 & 52.8 & 55.0 & 41.7 & 16.2 & 23.4 & 27.3 & 7.7 & 12.0 \\
Bin 1          & 33.4 & 13.7 & 19.4 & 42.3 & 39.9 & 41.0 & 18.2 & 8.1 & 11.2 & 24.4 & 5.7 & 9.2 \\
Bin 2          & 34.3 & 13.3 & 19.2 & 42.4 & 38.4 & 40.3 & 45.6 & 19.6 & 27.4 & 29.2 & 7.0 & 11.4 \\
Bin 3          & 32.6 & 14.7 & 20.3 & 46.4 & 40.9 & 43.5 & 43.8 & 15.4 & 22.8 & 25.0 & 7.2 & 11.1 \\
Bin 4          & 47.5 & 23.8 & 31.7 & 69.4 & 63.9 & 66.6 & 45.6 & 21.0 & 28.8 & 28.0 & 10.5 & 15.3 \\
\bottomrule
\end{tabular}%
}
\label{tab:baseline_LSTM}
\end{table*}

\begin{table*}
\centering
\footnotesize
\caption{Label Reasoning Network}
\resizebox{\textwidth}{!}{%
\begin{tabular}{l ccc ccc ccc ccc}
\toprule
 & \multicolumn{3}{c}{Overall} & \multicolumn{3}{c}{Coarse} & \multicolumn{3}{c}{Fine} & \multicolumn{3}{c}{Ultra-fine} \\
\cmidrule(lr){2-4} \cmidrule(lr){5-7} \cmidrule(lr){8-10} \cmidrule(lr){11-13}
Subset & P & R & F1 & P & R & F1 & P & R & F1 & P & R & F1 \\
\midrule
Full Test & 57.1 & 33.7 & 42.4 & 75.7 & 76.6 & 76.1 & 57.5 & 46.8 & 51.6 & 44.0 & 19.8 & 27.3 \\
Bin 1     & 52.7 & 27.4 & 36.0 & 65.8 & 69.5 & 67.6 & 52.8 & 42.1 & 46.9 & 43.4 & 14.5 & 21.8 \\
Bin 2     & 49.2 & 26.8 & 34.7 & 65.6 & 67.7 & 66.7 & 47.4 & 38.8 & 42.7 & 37.1 & 15.7 & 22.0 \\
Bin 3     & 52.3 & 27.9 & 36.4 & 67.3 & 68.7 & 68.0 & 55.0 & 37.6 & 44.7 & 38.8 & 17.3 & 23.9 \\
Bin 4     & 61.5 & 38.9 & 47.7 & 82.6 & 82.0 & 82.3 & 61.4 & 52.3 & 56.5 & 47.0 & 23.2 & 31.1 \\
\bottomrule
\end{tabular}%
}
\label{tab:LRN_seq}
\end{table*}


\begin{table*}
\centering
\footnotesize
\caption{Box4Types}
\resizebox{\textwidth}{!}{%
\begin{tabular}{l ccc ccc ccc ccc}
\toprule
Subset & \multicolumn{3}{c}{Overall} & \multicolumn{3}{c}{Coarse} & \multicolumn{3}{c}{Fine} & \multicolumn{3}{c}{Ultra-fine} \\
\cmidrule(lr){2-4} \cmidrule(lr){5-7} \cmidrule(lr){8-10} \cmidrule(lr){11-13}
           & P   & R   & F1  & P   & R   & F1  & P   & R   & F1  & P   & R   & F1 \\
\midrule
Full Test & 52.8 & 38.9 & 44.8 & 70.5 & 82.9 & 76.2 & 52.9 & 53.4 & 53.2 & 45.4 & 24.5 & 31.8 \\
Bin 1     & 46.3 & 31.0 & 37.2 & 58.9 & 77.7 & 67.0 & 42.9 & 43.6 & 43.2 & 38.8 & 16.0 & 22.7 \\
Bin 2     & 47.7 & 31.4 & 37.8 & 60.0 & 76.7 & 67.3 & 45.3 & 46.3 & 45.8 & 44.5 & 20.5 & 28.0 \\
Bin 3     & 47.8 & 33.9 & 39.7 & 60.2 & 79.9 & 68.7 & 49.6 & 53.8 & 51.6 & 42.8 & 20.6 & 27.8 \\
Bin 4     & 57.4 & 44.4 & 50.1 & 79.4 & 86.2 & 82.6 & 58.3 & 57.3 & 57.8 & 47.7 & 29.0 & 36.1 \\
\bottomrule
\end{tabular}%
}
\label{tab:box4types_box}
\end{table*}


\begin{table*}
\centering
\footnotesize
\caption{LITE}
\resizebox{\textwidth}{!}{%
\begin{tabular}{l ccc ccc ccc ccc}
\toprule
 & \multicolumn{3}{c}{Overall} & \multicolumn{3}{c}{Coarse} & \multicolumn{3}{c}{Fine} & \multicolumn{3}{c}{Ultra-fine} \\
\cmidrule(lr){2-4} \cmidrule(lr){5-7} \cmidrule(lr){8-10} \cmidrule(lr){11-13}
Subset & P & R & F1 & P & R & F1 & P & R & F1 & P & R & F1 \\
\midrule
Full Test  & 54.8 & 47.1 & 50.7 & 74.5 & 81.7 & 77.9 & 61.4 & 57.2 & 59.3 & 44.3 & 35.5 & 39.4 \\
Bin 1          & 50.7 & 41.4 & 45.6 & 64.7 & 76.3 & 70.0 & 53.8 & 42.4 & 47.4 & 41.0 & 31.8 & 35.8 \\
Bin 2          & 49.7 & 43.5 & 46.4 & 65.7 & 76.0 & 70.4 & 61.1 & 56.2 & 58.6 & 41.9 & 34.1 & 37.6 \\
Bin 3          & 49.2 & 45.6 & 47.3 & 63.9 & 73.9 & 68.6 & 57.1 & 53.7 & 55.4 & 43.3 & 36.3 & 39.5 \\
Bin 4          & 58.8 & 50.2 & 54.2 & 81.7 & 86.0 & 83.8 & 64.2 & 61.7 & 62.9 & 46.1 & 36.7 & 40.8 \\
\bottomrule
\end{tabular}%
}
\label{tab:LITE}
\end{table*}

\end{document}